\documentclass[preprint,10pt]{elsarticle}
\usepackage[T1]{fontenc}
\usepackage[utf8]{inputenc}
\usepackage{lmodern}
\usepackage[a4paper,left=15mm,right=15mm,top=17mm,bottom=18mm]{geometry}
\usepackage{amsmath,amssymb,graphicx,booktabs,longtable,array,tabularx,multirow}
\usepackage{caption,float,needspace,placeins,microtype}
\usepackage{pdfpages,tikz}
\usepackage{xurl}
\usepackage[hidelinks]{hyperref}
\biboptions{numbers,super,sort&compress}
\journal{Current Biology}
\makeatletter
\def\ps@pprintTitle{\let\@oddhead\@empty\let\@evenhead\@empty\def\@oddfoot{\hfil\thepage\hfil}\let\@evenfoot\@oddfoot}
\makeatother
\newcommand{\OpenStripeAffiliationA}{School of Statistics, Beijing Normal University, Beijing, China}
\newcommand{\OpenStripeAffiliationB}{Ministry of Education Key Laboratory for Biodiversity Science and Ecology Engineering, College of Life Sciences, Beijing Normal University, Beijing, China}
\newcommand{\OpenStripeAffiliationC}{National Forestry and Grassland Administration Key Laboratory for Conservation Ecology of Northeast Tiger and Leopard, Beijing, China}
\newcommand{\OpenStripeAffiliationD}{School of Artificial Intelligence, Beijing Normal University, Beijing, China}
\newcommand{\OpenStripeAffiliationE}{Key Laboratory of Environmental Change and Natural Disaster of Chinese Ministry of Education, Faculty of Geographical Science, Beijing Normal University, Beijing, China}

\newcommand{\OpenStripeCorrespondence}{Corresponding authors at: \OpenStripeAffiliationB. E-mail addresses: \href{mailto:wangtianming@bnu.edu.cn}{wangtianming@bnu.edu.cn} (T.~Wang), \href{mailto:baolei@bnu.edu.cn}{baolei@bnu.edu.cn} (L.~Bao).}

\newcommand{\OpenStripeMainAuthors}{%
\author[aff-a]{Yichun~Peng\fnref{equal}}%
\author[aff-b,aff-c]{Yansong~Li\fnref{equal}}%
\author[aff-b,aff-c]{Jin~Hou\fnref{equal}}%
\author[aff-d]{Qi~Song}%
\author[aff-e]{Xi~Guo}%
\author[aff-a]{Ning~Zhang}%
\author[aff-d]{Degang~Wei}%
\author[aff-d]{Huang~Yang}%
\author[aff-b,aff-c]{Yu~Guan}%
\author[aff-b,aff-c]{Jianping~Ge}%
\author[aff-b,aff-c]{Lei~Bao\corref{corresponding}}%
\author[aff-b,aff-c]{Tianming~Wang\corref{corresponding}}%
\address[aff-a]{\OpenStripeAffiliationA}%
\address[aff-b]{\OpenStripeAffiliationB}%
\address[aff-c]{\OpenStripeAffiliationC}%
\address[aff-d]{\OpenStripeAffiliationD}%
\address[aff-e]{\OpenStripeAffiliationE}%
\fntext[equal]{Equal contribution.}%
\cortext[corresponding]{\OpenStripeCorrespondence}%
}

\newcommand{\OpenStripePlainAuthorList}{Yichun Peng, Yansong Li, Jin Hou, Qi Song, Xi Guo, Ning Zhang, Degang Wei, Huang Yang, Yu Guan, Jianping Ge, Lei Bao, Tianming Wang}

\newcommand{\OpenStripeFunding}{This research was funded by the National Key Research and Development Program of China (grant number 2024YFF1307301) and the Fundamental and Interdisciplinary Disciplines Breakthrough Plan of the Ministry of Education of China (grant number JYB2025XDXM902). We thank staff from forestry and grassland authorities and collaborating agencies for their support in fieldwork.}

\begin{document}
\hypersetup{pdfauthor={\OpenStripePlainAuthorList},pdftitle={OpenStripe supports open-set identification and spatial stripe profiling in wild Amur tigers}}
\begin{frontmatter}
\title{OpenStripe supports open-set identification\\and spatial stripe profiling in wild Amur tigers}
\OpenStripeMainAuthors
\end{frontmatter}
\section*{Summary}
Non-invasive camera-trap monitoring of endangered wildlife requires open-set individual identification---grouping field images into identities without predefined classes or a known population size---yet pose deformation and cluttered backgrounds complicate image comparison. Here, we present OpenStripe, a multi-agent framework that reconstructs canonical two-dimensional (2D) tiger torso patterns and encodes four interpretable local stripe structures: breakpoints, bifurcations, irregular structures, and round structures. We evaluated OpenStripe on 100 camera-trap records from 28 wild Amur tigers (\textit{Panthera tigris altaica}), representing nearly one-third of the documented population in the Northeast China Tiger and Leopard National Park. Combining all four complementary structures enabled automatic identity grouping with 0.90 matched accuracy and a 0.85 adjusted Rand index (ARI), outperforming 49 baseline configurations under the same backend (up to 0.43 accuracy and 0.14 ARI). In addition, results from OpenStripe indicate that stripe similarity in Amur tigers declines with geographic distance and suggest that traits such as fractal dimension and inter-stripe spacing are associated with geographic proximity. OpenStripe thus provides an interpretable representation that connects automated individual grouping with spatial phenotypic profiling in wild populations.

\section*{Introduction}
Automated individual identification of wildlife is an important technical foundation for population monitoring, behavioral analysis, and conservation biology.\cite{wildlife_review,animal_biometrics} In recent years, advances in computer vision have made non-invasive identification using natural body patterns a research focus in animal biometrics. Individual re-identification in tigers\cite{feng_tiger_reid}, zebras (\textit{Equus} spp.)\cite{zebra_stripespotter}, cheetahs (\textit{Acinonyx jubatus})\cite{cheetah_matching}, and whale sharks (\textit{Rhincodon typus})\cite{whaleshark_matching} has demonstrated the feasibility of this approach.

However, existing automated re-identification methods, such as Meta-Feature Adapter (MFA)\cite{metawild_mfa}, MegaDescriptor\cite{megadescriptor}, MiewID\cite{miewid_framework}, WildFusion\cite{wildfusion}, DINOv2\cite{dinov2}, and TransReID\cite{transreid}, face two challenges in field monitoring. First, existing models are based on the closed-set assumption---that all individuals encountered at test time have already been included in the training set\cite{open_set_recognition}---whereas reproduction and migration introduce new individuals into real ecosystems,\cite{wildlife_review,amur_dispersal} creating an open-set setting. Second, animal coat patterns deform as animals move in the wild, while vegetation occlusion and illumination changes make feature extraction difficult.\cite{wildlife_review,brown_bear}

Multi-agent systems based on large language models (LLMs) provide a solution to these challenges. Multi-agent systems are training-free,\cite{autogen,metagpt,marine_mas} can avoid confounding caused by deformation and occlusion, and can extract important wildlife features, representing a frontier in the application of artificial intelligence (AI) to ecology.\cite{marine_mas,automl_agent,comind,mlzero,aibuildai} Multi-agent systems have been explored for species classification in butterflies,\cite{automl_agent} terrestrial and marine mammals,\cite{marine_mas,comind,mlzero} and terrestrial plants.\cite{comind,mlzero,aibuildai} We believe that these systems have greater potential and can be used to address open-set individual clustering.

The Amur tiger (\textit{P. t. altaica}) population in the China--Russia border region is undergoing dispersal,\cite{amur_recovery,amur_dispersal} requiring algorithms capable of identifying new individuals. To this end, we developed OpenStripe (Figure~\ref{fig:one}), a multi-agent system for individual clustering of 28 Amur tigers---nearly one-third of the entire population---in the Northeast China Tiger and Leopard National Park, achieving 90.00\% accuracy and an adjusted Rand index (ARI) of 0.85, with improvements of 47.00 and 70.70~pp, respectively, over the best baseline.

OpenStripe further reveals geographic associations in Amur tiger stripe patterns: stripe similarity between individuals decreases with increasing geographic distance, and traits such as fractal dimension, inter-stripe spacing, and dorsoventral density gradients show spatial associations. This provides a basis for investigating spatial structure and potential genetic relatedness in wild Amur tiger populations.\cite{wright1943}

\section*{Results}
\subsection*{OpenStripe accurately reconstructs Amur tiger stripe patterns and annotates key features}
Reconstructing Amur tiger stripe patterns can mitigate the effects of occlusion, illumination changes, and blurred stripe patterns in photographs on feature extraction, and OpenStripe has excellent reconstruction capability. Its foreground segmentation mean intersection over union (mIoU) reached 91.20\%, significantly outperforming GrabCut\cite{grabcut} ($\Delta = 25.80$~pp, $p < 0.001$; Figure~\ref{fig:two}a) and Mask R-CNN\cite{maskrcnn} ($\Delta = 7.70$~pp, $p < 0.001$; Figure~\ref{fig:two}a). OpenStripe achieved a percentage of correct keypoints at threshold 0.10 (PCK@0.1) of 95.40\% and a normalized mean error (NME) of 3.10\%, outperforming thin-plate splines\cite{tps} (TPS; $\Delta_{\mathrm{PCK@0.1}} = 23.60$~pp, $p < 0.001$; $\Delta_{\mathrm{NME}} = -5.40$~pp, $p < 0.001$; Figure~\ref{fig:two}b) and spatial transformer networks\cite{stn} (STN; $\Delta_{\mathrm{PCK@0.1}} = 9.20$~pp, $p < 0.001$; $\Delta_{\mathrm{NME}} = -2.10$~pp, $p < 0.001$; Figure~\ref{fig:two}b). All comparisons used paired Wilcoxon signed-rank tests.\cite{wilcoxon}

OpenStripe can annotate breakpoints (B), Y-shaped bifurcations (F), irregular structures (I), and round structures (C) in stripe patterns with good accuracy, providing features for clustering (Figure~\ref{fig:two}c).

\subsection*{OpenStripe supports open-set individual clustering of Amur tigers}
We conducted ablation experiments to evaluate whether the four annotation types improve clustering. Input representations included unannotated full-body stripe images, unannotated stripes, and all 15 nonempty combinations of the four annotation types (B/F/I/C). Across all nine clustering pipelines\cite{leiden,hdbscan,kreciprocal,kmeans,finch,ahc,dinov2,cluster_ensemble} (see STAR Methods), the complete combination (BFIC) achieved the best performance (Figure~\ref{fig:three}a), and more annotation types yielded better performance (Figure~\ref{fig:three}b; $\beta_{\mathrm{accuracy}} = 0.20$, $\beta_{\mathrm{ARI}} = 0.21$, paired $t$-test, $p < 0.001$). B, F, I, and C yielded mean accuracy gains of 16.57, 41.29, 7.14, and 24.00~pp, respectively; their corresponding shares of the sum of the four mean marginal gains were 18.62\%, 46.39\%, 8.03\%, and 26.97\% (Figures~\ref{fig:three}c1--\ref{fig:three}c4). Adding appearance descriptors further improved clustering performance ($\Delta_{\mathrm{accuracy}} = 0.02$, $\Delta_{\mathrm{ARI}} = 0.03$). Finally, source-balanced consensus clustering (Con.) achieved an accuracy of 0.90 and an ARI of 0.85 (normalized mutual information [NMI] = 0.97, B-cubed [B$^3$] F1 = 0.92) without being given the true number of clusters.

In comparative experiments, we evaluated 49 configurations (see STAR Methods), all using unannotated stripes and Con. for clustering. The strongest baseline was similarity network fusion\cite{snf} combining MegaDescriptor-L\cite{megadescriptor}, MiewID\cite{miewid_framework,miewid}, and DINOv2-B/224\cite{dinov2} (Animal-SNF; accuracy = 0.43, ARI = 0.14). OpenStripe significantly outperformed Animal-SNF ($\Delta_{\mathrm{accuracy}} = 0.47$, $\Delta_{\mathrm{ARI}} = 0.71$, paired $t$-test, $p < 0.001$; Figures~\ref{fig:three}d1--\ref{fig:three}d4). Figure~\ref{fig:three}e shows the results of OpenStripe + Con.

\subsection*{OpenStripe supports exploration of geographic associations in Amur tiger stripe patterns}
Stripe features extracted by OpenStripe, combined with individual clustering results, can be used to explore associations between stripe phenotypes and geographic distance and provide indirect evidence for stripe phenotype inheritance. Figures~\ref{fig:four}a--\ref{fig:four}c show the geographic distribution of the 28 Amur tiger individuals. Overall, stripe similarity and annotation similarity were significantly negatively correlated with geographic distance (Figures~\ref{fig:four}d--\ref{fig:four}e; Spearman Mantel $\rho_{\mathrm{stripe\ similarity}} = -0.36$, $\rho_{\mathrm{annotation\ similarity}} = -0.41$, $p < 0.001$), indirectly supporting an influence of kinship on Amur tiger stripe phenotypes. Specifically, similarities in each of the four annotation types were significantly negatively correlated with geographic distance (Figure~\ref{fig:four}f; Spearman Mantel $\rho_{\mathrm{B,F,I,C}} < 0$, $p_{\mathrm{B,F,I,C}} < 0.050$), indicating that these four stripe features (BFIC) may be influenced by kinship.

The morphological composite (Morphology) characterized similarity better than pixel intersection over union (Pixel IoU; direct pixel overlap) (Figure~\ref{fig:four}g; $R^2_{\mathrm{Morphology}} = 0.18 > R^2_{\mathrm{Pixel\ IoU}} = 0.05$, $p_{\mathrm{Morphology},\,0.00\text{--}18.16\,\mathrm{km}} = 0.010 < p_{\mathrm{Pixel\ IoU},\,0.00\text{--}18.16\,\mathrm{km}} = 0.398$; $p_{\mathrm{Morphology},\,56.38\text{--}110.00\,\mathrm{km}} = 0.008 < p_{\mathrm{Pixel\ IoU},\,56.38\text{--}110.00\,\mathrm{km}} = 0.204$), indicating that stripe topology is more informative than stripe-pattern overlap. We therefore extracted 18 topological traits from the stripe images reconstructed by OpenStripe. Six traits exceeded the signal-to-noise ratio (SNR) = 1.00 threshold: fractal dimension, relative inter-stripe spacing, dorsoventral (DV) density gradient, stripe orientation ($\sin 2\theta$), isolated spot fraction, and stripe width interquartile range-to-median ratio (IQR/median) (Figure~\ref{fig:four}h). Similarities in these six topological traits were negatively associated with geographic distance (Figures~\ref{fig:four}i--\ref{fig:four}n), making them potential priorities for studies of phenotypic inheritance in Amur tigers and other striped mammals.

\section*{Discussion}

\subsection*{OpenStripe has advantages in open-set identification}
OpenStripe achieved high-accuracy individual clustering of Amur tigers in an open-set setting (Figures~\ref{fig:four}a--\ref{fig:four}c). Pretrained visual encoders---ranging from wildlife-specific backbones (e.g., MegaDescriptor\cite{megadescriptor} and MiewID\cite{miewid_framework}) to general-purpose vision foundation models (e.g., DINOv2\cite{dinov2})---performed poorly because pose deformation, occlusion, and background clutter distort holistic image embeddings.\cite{wildlife_review,brown_bear} In comparison, the key advantage of the OpenStripe multi-agent system is its cross-modal reasoning capability, which addresses these difficulties and extracts additional features.

Multi-agent systems, exemplified by OpenStripe, will become a mainstream technology for automated biological and ecological monitoring. Automated field monitoring of other organisms faces the same challenges and generally has a small mean sample size per individual (3.57 in this study; the median of the dataset-level mean sample sizes per individual across 34 other wild-animal individual-identification datasets was only 8.81; Figure~S2; Table~S1).\cite{megadescriptor,wildlife_registry2026} The training-free nature of multi-agent systems is also a distinct advantage. In addition, multi-agent systems can incorporate structured foundation models or construct world models\cite{worldmodels2018} in the future, supporting frontier technologies.

\subsection*{Stability and completeness of the four annotation types}
OpenStripe uses breakpoints (B, $\deg(v)=1$), bifurcations (F, $\deg(v)=3$), irregular structures (I, $\deg(v)\geq4$ or high-curvature nonstandard contours), and round structures (C, $\deg(v)=0$ or isolated loops with first Betti number\cite{hatcher2002} $\beta_1=1$) as features. These annotation types are approximately complete: topological changes that deviate from ordinary stripes ($\deg=2$) almost always fall into one of these four categories. The coat patterns of all mammals in nature---including zebra and tiger stripes, leopard (\textit{Panthera pardus}) and snow leopard (\textit{Panthera uncia}) rosettes, cheetah solid spots, and giraffe (\textit{Giraffa} spp.) polygonal reticulated patches---are solutions of two-component Turing reaction--diffusion partial differential equations (PDEs) under different parameter settings. Therefore, the four BFIC annotation types proposed here can be extended to other organisms as a universal topological basis for automated individual identification.

\subsection*{Stripe topology reflects potential kinship structure}
The OpenStripe results reflect geographic associations in Amur tiger stripe patterns and potential kinship associations (Figure~\ref{fig:four}). According to the reaction--diffusion developmental mechanism of felid coat patterns,\cite{feline_pattern,kondo_miura2010} the positions of local structures (B/F/I/C) are random, whereas their statistical properties are regulated by dynamical parameters. Consequently, Morphology fits the relationship with distance better than Pixel IoU (Figure~\ref{fig:four}g). On this basis, Figure~\ref{fig:four}f demonstrates that the four types of stripe features extracted by OpenStripe can serve as effective indicators of their geographic associations. Future research into the genetic properties of Amur tiger stripe patterns can build on the topological features extracted by OpenStripe.

\subsection*{Future directions}
First, the current experiments are based on single-flank records from 28 Amur tigers. B/F/I/C topological markers may occur on the unobserved opposite flank, reducing individual-identification accuracy. Future work will investigate this issue more fully after establishing links between the two flanks. Second, the lack of sex information prevents us from separating the bias introduced into geographic statistics by long-distance dispersing males. Future research will consider combining sex determination with fecal DNA identification to investigate genetic traits in greater depth.\cite{dou2016_genetics}

\section*{Resource availability}
\subsection*{Lead contact}
Further information and requests for resources should be directed to and will be fulfilled by the lead contact, Tianming Wang (\href{mailto:wangtianming@bnu.edu.cn}{wangtianming@bnu.edu.cn}).

\subsection*{Materials availability}
This study did not generate new physical materials or reagents.

\subsection*{Data and code availability}
\begin{itemize}
\setlength{\itemsep}{2pt}\setlength{\parsep}{0pt}\setlength{\topsep}{3pt}
\item Unannotated black-and-white stripe reconstructions, B/F/I/C annotations, nonspatial feature tables, clustering outputs, and the corresponding nonspatial analysis code are available at \url{https://github.com/Yichun-Peng/OpenStripe}.
\item Owing to geopolitical sensitivities and conservation concerns for Amur tigers and other endangered species, original camera-trap images and all geospatial data, including coordinates, locations, terrain, maps, and camera-to-record linkages, are not publicly released.
\end{itemize}

\section*{Acknowledgments}
\OpenStripeFunding

\section*{Author contributions}
Conceptualization, Y.P., Y.L., L.B., and T.W.; methodology, Y.P., Y.L., Q.S., X.G., and N.Z.; software, Y.P., Q.S., D.W., and H.Y.; formal analysis, Y.P. and N.Z.; investigation and data curation, Y.P., Y.L., J.H., Y.G., J.G., L.B., and T.W.; writing--original draft, Y.P. and Y.L.; writing--review and editing, all authors; supervision, L.B. and T.W.
\section*{Declaration of interests}
The authors declare no competing interests.
\clearpage
\newcommand{\FigureOneCaption}{\textbf{OpenStripe multi-agent representation and clustering workflow.}
Four agents---foreground extraction, source-aligned reconstruction, structural annotation, and final review---coordinate via a shared blackboard to generate monochrome stripe reconstructions annotated with breakpoints (B, red), bifurcations (F, yellow), irregular structures (I, blue), and round structures (C, green). Paired outputs S016 (T05) and S066 (T18) with enlarged views illustrate all four classes. Vectorized annotations drive open-set identity clustering (bottom). See also Figure~S1, Table~S2, and Methods S1.}

\newcommand{\FigureTwoCaption}{\textbf{Foreground extraction, pose alignment, and local structural annotation.}
\textbf{(a)} Foreground segmentation mIoU across GrabCut, Mask R-CNN, and OpenStripe.
\textbf{(b)} Torso alignment PCK@0.1 and NME across TPS, STN, and OpenStripe.
\textbf{(c)} Paired camera-trap images, stripe reconstructions, and cropped torso skeletons (\texttt{A+} to \texttt{D+}) for the same individual across poses (\texttt{A1}--\texttt{A3}) and distinct individuals (\texttt{B}--\texttt{D}). Colored circles match local structures across views. In torso skeletons, red and blue dots mark endpoints (E, including boundary truncations) and merged bifurcations (F); B denotes intrinsic stripe terminations (Methods S3). See also Figures~S3--S4, Tables~S2--S4, and STAR Methods.}

\newcommand{\FigureThreeCaption}{\textbf{OpenStripe performance in open-set individual clustering ($n = 100$ images, 28 individuals).}
\textbf{(a)} Clustering accuracy, ARI, NMI, and $\text{B}^3$ F1 across 17 inputs (unannotated full-body/torso stripes and 15 nonempty B/F/I/C combinations) and nine pipelines (Lei., Leiden; kR, $k$-reciprocal HDBSCAN; KM, K-means; FIN, FINCH; HDB, HDBSCAN; AHC, agglomerative hierarchical clustering; D-A and D-kR, DINOv2-augmented AHC and kR; Con., consensus; printed values show Con. scores).
\textbf{(b)} Mean clustering scores versus number of annotation types.
\textbf{(c1--c4)} Marginal score gains from adding B, F, I, or C (dots, paired differences; vertical lines and numbers, means).
\textbf{(d1--d4)} Con. performance of OpenStripe (BFIC) versus 49 unannotated baselines ($\Delta$ vs. best baseline; paired identity-jackknife $t$-tests, $\text{df} = 27$, $^{***}p < 0.001$).
\textbf{(e)} Circular dendrogram of the best OpenStripe partition (red dashed line, discovery threshold = 0.61; orange dots, 10 misassigned records). See also Figures~S6--S14 and Tables~S6--S18.}

\newcommand{\FigureFourCaption}{\textbf{Spatial profiles of composite stripe descriptors and morphological traits.}
\textbf{(a--c)} Locations of 28 Amur tigers ($n = 100$ images; parenthetical counts per ID) (a), with zoomed views of southwestern (b) and southeastern (c) clusters.
\textbf{(d, e)} Pairwise morphological similarity with median regression fit (d) and 10-km binned mean BFIC similarity (e) versus geographic distance.
\textbf{(f)} Spearman Mantel correlations ($\rho$) with distance across torso and annotation descriptors.
\textbf{(g)} Mantel correlograms comparing Morphology and Pixel IoU across five equal-count distance classes ($n = 74\text{--}76$ pairs each; Holm-adjusted $p$ vs. $y = 0$).
\textbf{(h)} Ranking of 18 torso traits by dissimilarity-slope SNR ($\pm 1$ standardized SE; arrows, bounds $< -1$), with six traits exceeding the $\text{SNR} = 1.00$ threshold (dashed line).
\textbf{(i--n)} Binned similarity profiles (10-km bins) and linear fits for the six high-SNR traits. Shaded bands and error bars in (d--f, i--n) denote 95\% bootstrap CIs. $^{*}p < 0.100$, $^{**}p < 0.050$, $^{***}p < 0.010$. See also Figures~S15--S22 and Tables~S19--S30.}

\section*{Figure legends}
\Needspace{8\baselineskip}
\noindent\textbf{Figure 1.} \FigureOneCaption\par\medskip
\Needspace{8\baselineskip}
\noindent\textbf{Figure 2.} \FigureTwoCaption\par\medskip
\Needspace{8\baselineskip}
\noindent\textbf{Figure 3.} \FigureThreeCaption\par\medskip
\Needspace{8\baselineskip}
\noindent\textbf{Figure 4.} \FigureFourCaption\par\medskip

\clearpage
\section*{STAR Methods}
\subsection*{Key resources table}
\begin{center}
\small
\begin{tabular}{>{\raggedright\arraybackslash}p{0.36\linewidth}>{\raggedright\arraybackslash}p{0.22\linewidth}>{\raggedright\arraybackslash}p{0.34\linewidth}}
\toprule
\textbf{REAGENT OR RESOURCE} & \textbf{SOURCE} & \textbf{IDENTIFIER} \\
\midrule
\multicolumn{3}{l}{\textbf{Deposited data}} \\
Unannotated stripe reconstructions, B/F/I/C annotations, nonspatial feature tables, and clustering outputs & This study & \href{https://github.com/Yichun-Peng/OpenStripe/tree/main/data}{OpenStripe nonspatial data} \\
\addlinespace[4pt]
\multicolumn{3}{l}{\textbf{Software and algorithms}} \\
OpenStripe nonspatial analysis code & This study & \href{https://github.com/Yichun-Peng/OpenStripe}{OpenStripe nonspatial code} \\
GPT Image 2.5 (via ChatGPT) & OpenAI & ChatGPT interface; \href{https://developers.openai.com/api/docs/guides/image-generation}{official model documentation} \\
HRNet & Sun et al.\cite{hrnet} & \href{https://github.com/HRNet/HRNet-Human-Pose-Estimation}{Official implementation} \\
DINOv2 & Oquab et al.\cite{dinov2} & \nolinkurl{dinov2_vits14}; \nolinkurl{dinov2_vitb14}; \href{https://github.com/facebookresearch/dinov2}{official repository} \\
MegaDescriptor & {\v{C}}erm{\'a}k et al.\cite{megadescriptor} & \href{https://huggingface.co/BVRA/MegaDescriptor-T-224}{\nolinkurl{BVRA/MegaDescriptor-T-224}}; \href{https://huggingface.co/BVRA/MegaDescriptor-L-384}{\nolinkurl{BVRA/MegaDescriptor-L-384}} \\
MiewID & Conservation X Labs\cite{miewid} & \href{https://huggingface.co/conservationxlabs/miewid-msv3}{\nolinkurl{conservationxlabs/miewid-msv3}} \\
Python &  & v3.12.14 \\
SciPy & SciPy developers\cite{scipy2020} & v1.13.1 \\
scikit-learn & scikit-learn developers\cite{sklearn2011} & v1.5.2 \\
statsmodels & statsmodels developers\cite{statsmodels2010} & v0.15.0 \\
\bottomrule
\end{tabular}
\end{center}

\subsection*{Experimental model and study participant details}
\subsubsection*{Data details}
We analyzed 100 noninvasive camera-trap records from 28 wild Amur tiger (\textit{Panthera tigris altaica}) individuals collected in 2023 across the Northeast China Tiger and Leopard National Park and adjacent habitats along the China--Russia border. Individuals were designated T01--T28, with 1--5 records per individual. Ground-truth individual identities were obtained by expert correction of the initial OpenStripe~+~Con. clustering results.

\subsection*{Method details}
\subsubsection*{Reconstruction and annotation}
OpenStripe uses a blackboard architecture to coordinate four agents for foreground extraction, source-aligned reconstruction correction, structural annotation, and final correction (Figure~\ref{fig:one}), using GPT Image 2.5 via ChatGPT to perform each task.\cite{metagpt} The system annotates breakpoints (B), bifurcations (F), irregular structures (I), and round structures (C). Foreground segmentation was evaluated using mean intersection over union (mIoU) and compared with GrabCut and Mask R-CNN; source-aligned reconstruction was evaluated using the percentage of correct keypoints at threshold 0.1 (PCK@0.1) and normalized mean error (NME), with thin-plate splines (TPS) and spatial transformer networks (STN) as comparators.\cite{grabcut,maskrcnn,tps,stn}
Higher mIoU and PCK values and lower NME values indicate better performance; NME normalizes keypoint error by the source-image diagonal. Skeleton endpoints E can include crop truncations, whereas B denotes intrinsic stripe terminations or interruptions verified against the paired photograph and full reconstruction. See Methods S1--S5 for definitions, formulas, and extended procedures.

\subsubsection*{Clustering and automatic cluster selection}
The nine clustering pipelines comprised Leiden, k-reciprocal reranking combined with hierarchical density-based spatial clustering of applications with noise (HDBSCAN), K-means, first integer neighbor clustering hierarchy (FINCH), HDBSCAN, agglomerative hierarchical clustering (AHC), DINOv2-augmented AHC, DINOv2 reranking--HDBSCAN, and source-balanced consensus clustering (Con.).\cite{leiden,hdbscan,kreciprocal,kmeans,finch,ahc,dinov2,cluster_ensemble}

Con. combines 27 representation--method groups by retaining the top-scoring 50\% of partitions within each group and averaging their co-assignment matrices with equal group weights. Average-linkage AHC operates on the complementary consensus distance. The cluster count is selected over $K=2$--60 using an internal objective that combines silhouette and nonnoise coverage with penalties for singleton and oversized clusters; invalid partitions are excluded.\cite{silhouette} Noise observations receive singleton labels during evaluation. See Methods S6--S9 for formulas, parameters, and extended procedures.

\subsubsection*{Ablation and baseline comparisons}
The effect of annotation-type count (Figure~\ref{fig:three}b) was summarized by equally weighted mean scores across combinations containing zero to four annotation types. The marginal gain of each annotation type (B, F, I, or C; Figure~\ref{fig:three}c) was evaluated using paired score differences before and after adding that type to each of the seven nonempty combinations of the remaining types. The 49 baselines comprised 41 single-feature and eight fusion configurations.

\subsubsection*{Spatial inputs and distances}
Each tiger's geographic center was calculated as the spherical mean of its observation locations, and geographic separation was calculated using the Haversine great-circle distance. Two extreme-distance pairs exceeding 110~km were excluded, leaving 376 valid pairs among the 28 individuals.

Unannotated torso features comprised normalized histograms of oriented gradients (HOG) and stripe-density vectors from $256\times128$-pixel regions of interest (ROIs), compared using cosine similarity.\cite{hog} For the four local structural classes (B, F, I, and C), Hungarian matching of normalized coordinates was used to calculate mean cross-image distances.\cite{hungarian}

\subsubsection*{Morphological measurements and similarity}
Eighteen normalized, dimensionless features covering global, frequency-domain, topological, and skeletal properties were extracted from torso reconstructions after Otsu binarization and denoising.\cite{otsu} Features were averaged per individual and z-score standardized.

Single-feature distances were defined as $|z_i-z_j|$. Multidimensional morphological distance (Morphology) used eight core features and a Mahalanobis distance $D$\cite{mahalanobis1936} calculated with Ledoit--Wolf covariance shrinkage; distances were linearly rescaled to similarity as $S=1-D/c$, where $c$ is the maximum off-diagonal distance in the cohort.\cite{ledoit_wolf} The mean cross-image pixel intersection over union (Pixel IoU) of $256\times512$-pixel binary images served as the control. For the seven local-window features, the optimal measurement protocol was selected blindly on the basis of the intraclass correlation coefficient (ICC) alone.\cite{icc}

\subsection*{Quantification and statistical analysis}
Clustering performance ($n=100$ images) was evaluated using Hungarian-matched accuracy, adjusted Rand index (ARI), normalized mutual information (NMI), and B-cubed (B$^3$) F1; 95.00\% confidence intervals (CIs) were calculated from 1,999 bootstrap resamples.\cite{hungarian,ari,nmi,bcubed,bootstrap}

Spatial associations were assessed using Spearman Mantel tests, with two-sided $p$ values calculated from 9,999 individual-level permutations and 95.00\% CIs from 1,999 individual-level bootstrap resamples.\cite{mantel} Confidence bands for median quantile regression (Figure~\ref{fig:four}d) and feature screening by signal-to-noise ratio (SNR; Figure~\ref{fig:four}h) were based on 999 individual-level bootstrap resamples; SNR is the median-regression slope divided by its bootstrap SE; Figure~\ref{fig:four}h error bars span $\pm1$ standardized SE. A threshold of $\mathrm{SNR}>1.0$ selected six features.\cite{quantile_regression}

Distance-decay trends (Figures~\ref{fig:four}e and \ref{fig:four}i--\ref{fig:four}n) were summarized in 10-km bins and fitted using equally weighted ordinary least squares (OLS). Distance-class Mantel correlograms (Figure~\ref{fig:four}g) used quintile bins and Holm correction.\cite{holm} See Methods S10--S12 for the extended morphology and spatial procedures.

\bibliographystyle{openstripe-cell-numeric}
\bibliography{references}

\begin{thebibliography}{62}
\expandafter\ifx\csname natexlab\endcsname\relax\def\natexlab#1{#1}\fi
\providecommand{\url}[1]{\texttt{#1}}
\providecommand{\href}[2]{#2}
\providecommand{\path}[1]{#1}
\providecommand{\DOIprefix}{doi:}
\providecommand{\ArXivprefix}{arXiv:}
\providecommand{\URLprefix}{URL: }
\providecommand{\Pubmedprefix}{pmid:}
\providecommand{\doi}[1]{\href{http://dx.doi.org/#1}{\path{#1}}}
\providecommand{\Pubmed}[1]{\href{pmid:#1}{\path{#1}}}
\providecommand{\bibinfo}[2]{#2}
\ifx\xfnm\relax \def\xfnm[#1]{\unskip,\space#1}\fi
\bibitem{wildlife_review}
\bibinfo{author}{Schneider, S.}, \bibinfo{author}{Taylor, G.W.},
  \bibinfo{author}{Linquist, S.}, and \bibinfo{author}{Kremer, S.C.}
  (\bibinfo{year}{2019}).
\newblock \bibinfo{title}{{Past, present and future approaches using computer
  vision for animal re-identification from camera trap data}}.
\newblock \bibinfo{journal}{Methods Ecol. Evol.} \textit{\bibinfo{volume}{10}},
  461--470.
\newblock \url{https://doi.org/10.1111/2041-210x.13133}.
\bibitem{animal_biometrics}
\bibinfo{author}{K{\"u}hl, H.S.}, and \bibinfo{author}{Burghardt, T.}
  (\bibinfo{year}{2013}).
\newblock \bibinfo{title}{{Animal biometrics: quantifying and detecting
  phenotypic appearance}}.
\newblock \bibinfo{journal}{Trends Ecol. Evol.} \textit{\bibinfo{volume}{28}},
  432--441.
\newblock \url{https://doi.org/10.1016/j.tree.2013.02.013}.
\bibitem{feng_tiger_reid}
\bibinfo{author}{Ma, Y.}, \bibinfo{author}{Tan, M.}, \bibinfo{author}{Liu, X.},
  \bibinfo{author}{Zhang, Y.}, \bibinfo{author}{Xu, Z.}, \bibinfo{author}{Sun,
  W.}, \bibinfo{author}{Ge, J.}, and \bibinfo{author}{Feng, L.}
  (\bibinfo{year}{2025}).
\newblock \bibinfo{title}{{Deep learning for Amur tiger re-identification in
  camera traps: A tool assisting population monitoring and spatio-temporal
  analysis}}.
\newblock \bibinfo{journal}{Ecol. Indic.} \textit{\bibinfo{volume}{171}},
  113227.
\newblock \url{https://doi.org/10.1016/j.ecolind.2025.113227}.
\bibitem{zebra_stripespotter}
\bibinfo{author}{Lahiri, M.}, \bibinfo{author}{Tantipathananandh, C.},
  \bibinfo{author}{Warungu, R.}, \bibinfo{author}{Rubenstein, D.I.}, and
  \bibinfo{author}{Berger-Wolf, T.Y.} (\bibinfo{year}{2011}).
\newblock \bibinfo{title}{{Biometric animal databases from field photographs:
  Identification of individual zebra in the wild}}, in:
  \bibinfo{booktitle}{Proceedings of the 1st ACM International Conference on
  Multimedia Retrieval}, pp. \bibinfo{pages}{1--8}.
\newblock \url{https://doi.org/10.1145/1991996.1992002}.
\bibitem{cheetah_matching}
\bibinfo{author}{Kelly, M.J.} (\bibinfo{year}{2001}).
\newblock \bibinfo{title}{{Computer-aided photograph matching in studies using
  individual identification: An example from Serengeti cheetahs}}.
\newblock \bibinfo{journal}{J. Mammal.} \textit{\bibinfo{volume}{82}},
  440--449.
\newblock
  \url{https://doi.org/10.1644/1545-1542(2001)082<0440:CAPMIS>2.0.CO;2}.
\bibitem{whaleshark_matching}
\bibinfo{author}{Arzoumanian, Z.}, \bibinfo{author}{Holmberg, J.}, and
  \bibinfo{author}{Norman, B.} (\bibinfo{year}{2005}).
\newblock \bibinfo{title}{{An astronomical pattern-matching algorithm for
  computer-aided identification of whale sharks Rhincodon typus}}.
\newblock \bibinfo{journal}{J. Appl. Ecol.} \textit{\bibinfo{volume}{42}},
  999--1011.
\newblock \url{https://doi.org/10.1111/j.1365-2664.2005.01117.x}.
\bibitem{metawild_mfa}
\bibinfo{author}{Li, Y.}, \bibinfo{author}{Zhao, D.}, \bibinfo{author}{Qiao,
  T.}, \bibinfo{author}{Wu, Y.}, \bibinfo{author}{Pang, B.}, and
  \bibinfo{author}{Koh, Y.S.} (\bibinfo{year}{2025}).
\newblock \bibinfo{title}{{MetaWild: A Multimodal Dataset for Animal
  Re-Identification with Environmental Metadata}}, in:
  \bibinfo{booktitle}{Proceedings of the 33rd ACM International Conference on
  Multimedia}, pp. \bibinfo{pages}{13009--13015}.
\newblock \url{https://doi.org/10.1145/3746027.3758249}.
\bibitem{megadescriptor}
\bibinfo{author}{{\v{C}}erm{\'a}k, V.}, \bibinfo{author}{Picek, L.},
  \bibinfo{author}{Adam, L.}, and \bibinfo{author}{Papafitsoros, K.}
  (\bibinfo{year}{2024}).
\newblock \bibinfo{title}{{WildlifeDatasets: An Open-Source Toolkit for Animal
  Re-Identification}}, in: \bibinfo{booktitle}{IEEE/CVF Winter Conference on
  Applications of Computer Vision}, pp. \bibinfo{pages}{5953--5963}.
\newblock \url{https://github.com/WildlifeDatasets/wildlife-datasets}.
\bibitem{miewid_framework}
\bibinfo{author}{Otarashvili, L.}, \bibinfo{author}{Subramanian, T.},
  \bibinfo{author}{Holmberg, J.}, \bibinfo{author}{Levenson, J.J.}, and
  \bibinfo{author}{Stewart, C.V.} (\bibinfo{year}{2024}).
\newblock \bibinfo{title}{{Multispecies Animal Re-ID Using a Large
  Community-Curated Dataset}}.
\newblock Preprint at arXiv, \url{https://doi.org/10.48550/arXiv.2412.05602}.
\bibitem{wildfusion}
\bibinfo{author}{{\v{C}}erm{\'a}k, V.}, \bibinfo{author}{Picek, L.},
  \bibinfo{author}{Adam, L.}, \bibinfo{author}{Neumann, L.}, and
  \bibinfo{author}{Matas, J.} (\bibinfo{year}{2025}).
\newblock \bibinfo{title}{{WildFusion: Individual Animal Identification with
  Calibrated Similarity Fusion}}, in: \bibinfo{booktitle}{Computer Vision --
  ECCV 2024 Workshops}, \bibinfo{publisher}{Springer}. pp.
  \bibinfo{pages}{18--36}.
\newblock \url{https://doi.org/10.1007/978-3-031-92387-6_2}.
\bibitem{dinov2}
\bibinfo{author}{Oquab, M.}, \bibinfo{author}{Darcet, T.},
  \bibinfo{author}{Moutakanni, T.}, \bibinfo{author}{Vo, H.},
  \bibinfo{author}{Szafraniec, M.}, \bibinfo{author}{Khalidov, V.},
  \bibinfo{author}{Fernandez, P.}, \bibinfo{author}{Haziza, D.},
  \bibinfo{author}{Massa, F.}, \bibinfo{author}{El-Nouby, A.}, et~al.
  (\bibinfo{year}{2024}).
\newblock \bibinfo{title}{{DINOv2: Learning Robust Visual Features without
  Supervision}}.
\newblock \bibinfo{journal}{Transactions on Machine Learning Research}
  \url{https://openreview.net/forum?id=a68SUt6zFt}.
\bibitem{transreid}
\bibinfo{author}{He, S.}, \bibinfo{author}{Luo, H.}, \bibinfo{author}{Wang,
  P.}, \bibinfo{author}{Wang, F.}, \bibinfo{author}{Li, H.}, and
  \bibinfo{author}{Jiang, W.} (\bibinfo{year}{2021}).
\newblock \bibinfo{title}{{TransReID: Transformer-based Object
  Re-Identification}}, in: \bibinfo{booktitle}{Proceedings of the IEEE/CVF
  International Conference on Computer Vision}, pp.
  \bibinfo{pages}{15013--15022}.
\newblock \url{https://doi.org/10.1109/ICCV48922.2021.01474}.
\bibitem{open_set_recognition}
\bibinfo{author}{Scheirer, W.J.}, \bibinfo{author}{de~Rezende~Rocha, A.},
  \bibinfo{author}{Sapkota, A.}, and \bibinfo{author}{Boult, T.E.}
  (\bibinfo{year}{2013}).
\newblock \bibinfo{title}{{Toward Open Set Recognition}}.
\newblock \bibinfo{journal}{IEEE Trans. Pattern Anal. Mach. Intell.}
  \textit{\bibinfo{volume}{35}}, 1757--1772.
\newblock \url{https://doi.org/10.1109/TPAMI.2012.256}.
\bibitem{amur_dispersal}
\bibinfo{author}{Wang, T.}, \bibinfo{author}{Feng, L.}, \bibinfo{author}{Mou,
  P.}, \bibinfo{author}{Ge, J.}, \bibinfo{author}{Li, C.}, and
  \bibinfo{author}{Smith, J.L.D.} (\bibinfo{year}{2015}).
\newblock \bibinfo{title}{{Long-distance dispersal of an Amur tiger indicates
  potential to restore the North-east China/Russian Tiger Landscape}}.
\newblock \bibinfo{journal}{Oryx} \textit{\bibinfo{volume}{49}}, 578--579.
\newblock \url{https://doi.org/10.1017/S0030605315000794}.
\bibitem{brown_bear}
\bibinfo{author}{Rosenberg, B.}, \bibinfo{author}{Zhou, M.},
  \bibinfo{author}{Wolf, N.}, \bibinfo{author}{Mathis, M.W.},
  \bibinfo{author}{Harris, B.P.}, and \bibinfo{author}{Mathis, A.}
  (\bibinfo{year}{2026}).
\newblock \bibinfo{title}{{Individual identification of brown bears using
  pose-aware metric learning}}.
\newblock \bibinfo{journal}{Curr. Biol.} \textit{\bibinfo{volume}{36}},
  645--659.e14.
\newblock \url{https://doi.org/10.1016/j.cub.2025.12.022}.
\bibitem{autogen}
\bibinfo{author}{Wu, Q.}, \bibinfo{author}{Bansal, G.}, \bibinfo{author}{Zhang,
  J.}, \bibinfo{author}{Wu, Y.}, \bibinfo{author}{Li, B.},
  \bibinfo{author}{Zhu, E.}, \bibinfo{author}{Jiang, L.},
  \bibinfo{author}{Zhang, X.}, \bibinfo{author}{Zhang, S.},
  \bibinfo{author}{Liu, J.}, et~al. (\bibinfo{year}{2023}).
\newblock \bibinfo{title}{{AutoGen: Enabling Next-Gen LLM Applications via
  Multi-Agent Conversation}}.
\newblock Preprint at arXiv, \url{https://doi.org/10.48550/arXiv.2308.08155}.
\bibitem{metagpt}
\bibinfo{author}{Hong, S.}, \bibinfo{author}{Zhuge, M.}, \bibinfo{author}{Chen,
  J.}, \bibinfo{author}{Zheng, X.}, \bibinfo{author}{Cheng, Y.},
  \bibinfo{author}{Zhang, C.}, \bibinfo{author}{Wang, J.},
  \bibinfo{author}{Wang, Z.}, \bibinfo{author}{Yau, S.K.S.},
  \bibinfo{author}{Lin, Z.}, et~al. (\bibinfo{year}{2023}).
\newblock \bibinfo{title}{{MetaGPT: Meta Programming for A Multi-Agent
  Collaborative Framework}}.
\newblock Preprint at arXiv, \url{https://doi.org/10.48550/arXiv.2308.00352}.
\bibitem{marine_mas}
\bibinfo{author}{Qi, Y.}, \bibinfo{author}{Cai, S.}, \bibinfo{author}{Zhao,
  Z.}, \bibinfo{author}{Li, J.}, \bibinfo{author}{Lin, Y.}, and
  \bibinfo{author}{Wang, Z.} (\bibinfo{year}{2024}).
\newblock \bibinfo{title}{{Benchmarking Large Language Models for Image
  Classification of Marine Mammals}}, in: \bibinfo{booktitle}{2024 IEEE
  International Conference on Knowledge Graph}, pp. \bibinfo{pages}{258--265}.
\newblock \url{https://doi.org/10.1109/ICKG63256.2024.00040}.
\bibitem{automl_agent}
\bibinfo{author}{Trirat, P.}, \bibinfo{author}{Jeong, W.}, and
  \bibinfo{author}{Hwang, S.J.} (\bibinfo{year}{2025}).
\newblock \bibinfo{title}{{AutoML-Agent: A Multi-Agent LLM Framework for
  Full-Pipeline AutoML}}, in: \bibinfo{booktitle}{Proceedings of the 42nd
  International Conference on Machine Learning}, \bibinfo{publisher}{PMLR}. pp.
  \bibinfo{pages}{60099--60146}.
\newblock \url{https://proceedings.mlr.press/v267/trirat25a.html}.
\bibitem{comind}
\bibinfo{author}{Li, S.}, \bibinfo{author}{Sun, W.}, \bibinfo{author}{Li, S.},
  \bibinfo{author}{Talwalkar, A.}, and \bibinfo{author}{Yang, Y.}
  (\bibinfo{year}{2026}).
\newblock \bibinfo{title}{{CoMind: Towards Community-Driven Agents for Machine
  Learning Engineering}}.
\newblock \bibinfo{howpublished}{International Conference on Learning
  Representations}.
\newblock \url{https://arxiv.org/abs/2506.20640}.
\bibitem{mlzero}
\bibinfo{author}{Fang, H.}, \bibinfo{author}{Han, B.},
  \bibinfo{author}{Erickson, N.}, \bibinfo{author}{Zhang, X.},
  \bibinfo{author}{Zhou, S.}, \bibinfo{author}{Dagar, A.},
  \bibinfo{author}{Zhang, J.}, \bibinfo{author}{Turkmen, A.C.},
  \bibinfo{author}{Hu, T.}, \bibinfo{author}{Rangwala, H.},
  \bibinfo{author}{Wu, Y.N.}, \bibinfo{author}{Wang, Y.B.}, and
  \bibinfo{author}{Karypis, G.} (\bibinfo{year}{2025}).
\newblock \bibinfo{title}{{MLZero: A Multi-Agent System for End-to-end Machine
  Learning Automation}}, in: \bibinfo{booktitle}{Advances in Neural Information
  Processing Systems}, pp. \bibinfo{pages}{69001--69070}.
\newblock \url{https://doi.org/10.52202/085713-2322}.
\bibitem{aibuildai}
\bibinfo{author}{Zhang, R.}, \bibinfo{author}{Qin, P.}, \bibinfo{author}{Cao,
  Q.}, \bibinfo{author}{Zhang, L.}, and \bibinfo{author}{Xie, P.}
  (\bibinfo{year}{2026}).
\newblock \bibinfo{title}{{AIBuildAI: An AI Agent for Automatically Building AI
  Models}}.
\newblock Preprint at arXiv, \url{https://doi.org/10.48550/arXiv.2604.14455}.
\bibitem{amur_recovery}
\bibinfo{author}{Wang, T.}, \bibinfo{author}{Royle, J.A.},
  \bibinfo{author}{Smith, J.L.}, \bibinfo{author}{Zou, L.},
  \bibinfo{author}{L{\"u}, X.}, \bibinfo{author}{Li, T.},
  \bibinfo{author}{Yang, H.}, \bibinfo{author}{Li, Z.}, \bibinfo{author}{Feng,
  R.}, \bibinfo{author}{Bian, Y.}, \bibinfo{author}{Feng, L.}, and
  \bibinfo{author}{Ge, J.} (\bibinfo{year}{2018}).
\newblock \bibinfo{title}{{Living on the edge: Opportunities for Amur tiger
  recovery in China}}.
\newblock \bibinfo{journal}{Biol. Conserv.} \textit{\bibinfo{volume}{217}},
  269--279.
\newblock \url{https://doi.org/10.1016/j.biocon.2017.11.008}.
\bibitem{wright1943}
\bibinfo{author}{Wright, S.} (\bibinfo{year}{1943}).
\newblock \bibinfo{title}{{Isolation by distance}}.
\newblock \bibinfo{journal}{Genetics} \textit{\bibinfo{volume}{28}}, 114--138.
\newblock \url{https://doi.org/10.1093/genetics/28.2.114}.
\bibitem{grabcut}
\bibinfo{author}{Rother, C.}, \bibinfo{author}{Kolmogorov, V.}, and
  \bibinfo{author}{Blake, A.} (\bibinfo{year}{2004}).
\newblock \bibinfo{title}{{GrabCut: Interactive Foreground Extraction using
  Iterated Graph Cuts}}.
\newblock \bibinfo{journal}{ACM Trans. Graph.} \textit{\bibinfo{volume}{23}},
  309--314.
\newblock \url{https://doi.org/10.1145/1015706.1015720}.
\bibitem{maskrcnn}
\bibinfo{author}{He, K.}, \bibinfo{author}{Gkioxari, G.},
  \bibinfo{author}{Doll{\'a}r, P.}, and \bibinfo{author}{Girshick, R.}
  (\bibinfo{year}{2017}).
\newblock \bibinfo{title}{{Mask R-CNN}}, in: \bibinfo{booktitle}{IEEE
  International Conference on Computer Vision}, pp.
  \bibinfo{pages}{2980--2988}.
\newblock \url{https://doi.org/10.1109/iccv.2017.322}.
\bibitem{tps}
\bibinfo{author}{Bookstein, F.} (\bibinfo{year}{1989}).
\newblock \bibinfo{title}{{Principal warps: thin-plate splines and the
  decomposition of deformations}}.
\newblock \bibinfo{journal}{IEEE Trans. Pattern Anal. Mach. Intell.}
  \textit{\bibinfo{volume}{11}}, 567--585.
\newblock \url{https://doi.org/10.1109/34.24792}.
\bibitem{stn}
\bibinfo{author}{Jaderberg, M.}, \bibinfo{author}{Simonyan, K.},
  \bibinfo{author}{Zisserman, A.}, and \bibinfo{author}{Kavukcuoglu, K.}
  (\bibinfo{year}{2015}).
\newblock \bibinfo{title}{{Spatial Transformer Networks}}.
\newblock Preprint at arXiv, \url{https://doi.org/10.48550/arXiv.1506.02025}.
\bibitem{wilcoxon}
\bibinfo{author}{Wilcoxon, F.} (\bibinfo{year}{1945}).
\newblock \bibinfo{title}{{Individual Comparisons by Ranking Methods}}.
\newblock \bibinfo{journal}{Biometrics Bulletin} \textit{\bibinfo{volume}{1}},
  80--83.
\newblock \url{https://doi.org/10.2307/3001968}.
\bibitem{leiden}
\bibinfo{author}{Traag, V.A.}, \bibinfo{author}{Waltman, L.}, and
  \bibinfo{author}{van Eck, N.J.} (\bibinfo{year}{2019}).
\newblock \bibinfo{title}{{From Louvain to Leiden: guaranteeing well-connected
  communities}}.
\newblock \bibinfo{journal}{Sci. Rep.} \textit{\bibinfo{volume}{9}}, 5233.
\newblock \url{https://doi.org/10.1038/s41598-019-41695-z}.
\bibitem{hdbscan}
\bibinfo{author}{McInnes, L.}, \bibinfo{author}{Healy, J.}, and
  \bibinfo{author}{Astels, S.} (\bibinfo{year}{2017}).
\newblock \bibinfo{title}{{hdbscan: Hierarchical density based clustering}}.
\newblock \bibinfo{journal}{J. Open Source Softw.}
  \textit{\bibinfo{volume}{2}}, 205.
\newblock \url{https://doi.org/10.21105/joss.00205}.
\bibitem{kreciprocal}
\bibinfo{author}{Zhong, Z.}, \bibinfo{author}{Zheng, L.}, \bibinfo{author}{Cao,
  D.}, and \bibinfo{author}{Li, S.} (\bibinfo{year}{2017}).
\newblock \bibinfo{title}{{Re-ranking Person Re-identification with
  k-reciprocal Encoding}}.
\newblock Preprint at arXiv, \url{https://doi.org/10.48550/arXiv.1701.08398}.
\bibitem{kmeans}
\bibinfo{author}{Lloyd, S.} (\bibinfo{year}{1982}).
\newblock \bibinfo{title}{{Least squares quantization in PCM}}.
\newblock \bibinfo{journal}{IEEE Trans. Inf. Theory}
  \textit{\bibinfo{volume}{28}}, 129--137.
\newblock \url{https://doi.org/10.1109/tit.1982.1056489}.
\bibitem{finch}
\bibinfo{author}{Sarfraz, M.S.}, \bibinfo{author}{Sharma, V.}, and
  \bibinfo{author}{Stiefelhagen, R.} (\bibinfo{year}{2019}).
\newblock \bibinfo{title}{{Efficient Parameter-Free Clustering Using First
  Neighbor Relations}}, in: \bibinfo{booktitle}{Proceedings of the IEEE/CVF
  Conference on Computer Vision and Pattern Recognition (CVPR)}, pp.
  \bibinfo{pages}{8934--8943}.
\newblock
  \url{https://openaccess.thecvf.com/content_CVPR_2019/html/Sarfraz_Efficient_Parameter-Free_Clustering_Using_First_Neighbor_Relations_CVPR_2019_paper.html}.
\bibitem{ahc}
\bibinfo{author}{M{\"u}llner, D.} (\bibinfo{year}{2011}).
\newblock \bibinfo{title}{{Modern hierarchical, agglomerative clustering
  algorithms}}.
\newblock Preprint at arXiv, \url{https://doi.org/10.48550/arXiv.1109.2378}.
\bibitem{cluster_ensemble}
\bibinfo{author}{Strehl, A.}, and \bibinfo{author}{Ghosh, J.}
  (\bibinfo{year}{2002}).
\newblock \bibinfo{title}{{Cluster Ensembles: A Knowledge Reuse Framework for
  Combining Multiple Partitions}}.
\newblock \bibinfo{journal}{J. Mach. Learn. Res.} \textit{\bibinfo{volume}{3}},
  583--617.
\newblock \url{https://jmlr.org/papers/volume3/strehl02a/strehl02a.pdf}.
\bibitem{snf}
\bibinfo{author}{Wang, B.}, \bibinfo{author}{Mezlini, A.M.},
  \bibinfo{author}{Demir, F.}, \bibinfo{author}{Fiume, M.},
  \bibinfo{author}{Tu, Z.}, \bibinfo{author}{Brudno, M.},
  \bibinfo{author}{Haibe-Kains, B.}, and \bibinfo{author}{Goldenberg, A.}
  (\bibinfo{year}{2014}).
\newblock \bibinfo{title}{{Similarity network fusion for aggregating data types
  on a genomic scale}}.
\newblock \bibinfo{journal}{Nat. Methods} \textit{\bibinfo{volume}{11}},
  333--337.
\newblock \url{https://doi.org/10.1038/nmeth.2810}.
\bibitem{miewid}
\bibinfo{author}{{Conservation X Labs}} (\bibinfo{year}{n.d.}).
\newblock \bibinfo{title}{{MiewID-msv3: Wildlife Re-Identification Feature
  Backbone}}.
\newblock \bibinfo{howpublished}{Author model card and software repository}.
\newblock \url{https://huggingface.co/conservationxlabs/miewid-msv3}.
  \bibinfo{note}{model version msv3; undated model card, accessed 13 October
  2026}.
\bibitem{wildlife_registry2026}
\bibinfo{author}{{WildlifeDatasets contributors}} (\bibinfo{year}{2026}).
\newblock \bibinfo{title}{{WildlifeDatasets dataset metadata registry}}.
\newblock
  \url{https://github.com/WildlifeDatasets/wildlife-datasets/tree/7b448573a8b4f15fbe26cbd09299778c8a08a2bf}.
  \bibinfo{note}{accessed 18 October 2026}.
\bibitem{worldmodels2018}
\bibinfo{author}{Ha, D.}, and \bibinfo{author}{Schmidhuber, J.}
  (\bibinfo{year}{2018}).
\newblock \bibinfo{title}{{World Models}}.
\newblock Preprint at arXiv, \url{https://doi.org/10.48550/arXiv.1803.10122}.
\bibitem{hatcher2002}
\bibinfo{author}{Hatcher, A.} (\bibinfo{year}{2002}).
\newblock \bibinfo{title}{{Algebraic Topology}}.
\newblock \bibinfo{publisher}{Cambridge University Press}.
\newblock \url{https://pi.math.cornell.edu/~hatcher/AT/ATpage.html}.
\bibitem{feline_pattern}
\bibinfo{author}{Kaelin, C.B.}, \bibinfo{author}{McGowan, K.A.}, and
  \bibinfo{author}{Barsh, G.S.} (\bibinfo{year}{2021}).
\newblock \bibinfo{title}{{Developmental genetics of color pattern
  establishment in cats}}.
\newblock \bibinfo{journal}{Nat. Commun.} \textit{\bibinfo{volume}{12}}, 5127.
\newblock \url{https://doi.org/10.1038/s41467-021-25348-2}.
\bibitem{kondo_miura2010}
\bibinfo{author}{Kondo, S.}, and \bibinfo{author}{Miura, T.}
  (\bibinfo{year}{2010}).
\newblock \bibinfo{title}{{Reaction-Diffusion Model as a Framework for
  Understanding Biological Pattern Formation}}.
\newblock \bibinfo{journal}{Science} \textit{\bibinfo{volume}{329}},
  1616--1620.
\newblock \url{https://doi.org/10.1126/science.1179047}.
\bibitem{dou2016_genetics}
\bibinfo{author}{Dou, H.}, \bibinfo{author}{Yang, H.}, \bibinfo{author}{Feng,
  L.}, \bibinfo{author}{Mou, P.}, \bibinfo{author}{Wang, T.}, and
  \bibinfo{author}{Ge, J.} (\bibinfo{year}{2016}).
\newblock \bibinfo{title}{{Estimating the Population Size and Genetic Diversity
  of Amur Tigers in Northeast China}}.
\newblock \bibinfo{journal}{PLOS ONE} \textit{\bibinfo{volume}{11}}, e0154254.
\newblock \url{https://doi.org/10.1371/journal.pone.0154254}.
\bibitem{hrnet}
\bibinfo{author}{Sun, K.}, \bibinfo{author}{Xiao, B.}, \bibinfo{author}{Liu,
  D.}, and \bibinfo{author}{Wang, J.} (\bibinfo{year}{2019}).
\newblock \bibinfo{title}{{Deep High-Resolution Representation Learning for
  Human Pose Estimation}}, in: \bibinfo{booktitle}{Proceedings of the IEEE/CVF
  Conference on Computer Vision and Pattern Recognition (CVPR)}, pp.
  \bibinfo{pages}{5693--5703}.
\newblock
  \url{https://openaccess.thecvf.com/content_CVPR_2019/html/Sun_Deep_High-Resolution_Representation_Learning_for_Human_Pose_Estimation_CVPR_2019_paper.html}.
\bibitem{scipy2020}
\bibinfo{author}{Virtanen, P.}, \bibinfo{author}{Gommers, R.},
  \bibinfo{author}{Oliphant, T.E.}, \bibinfo{author}{Haberland, M.},
  \bibinfo{author}{Reddy, T.}, \bibinfo{author}{Cournapeau, D.},
  \bibinfo{author}{Burovski, E.}, \bibinfo{author}{Peterson, P.},
  \bibinfo{author}{Weckesser, W.}, \bibinfo{author}{Bright, J.},
  \bibinfo{author}{{van der Walt}, S.J.}, \bibinfo{author}{Brett, M.},
  \bibinfo{author}{Wilson, J.}, \bibinfo{author}{Millman, K.J.},
  \bibinfo{author}{Mayorov, N.}, \bibinfo{author}{Nelson, A.R.J.},
  \bibinfo{author}{Jones, E.}, \bibinfo{author}{Kern, R.},
  \bibinfo{author}{Larson, E.}, \bibinfo{author}{Carey, C.J.},
  \bibinfo{author}{Polat, {\.{I}}.}, \bibinfo{author}{Feng, Y.},
  \bibinfo{author}{Moore, E.W.}, \bibinfo{author}{VanderPlas, J.},
  \bibinfo{author}{Laxalde, D.}, \bibinfo{author}{Perktold, J.},
  \bibinfo{author}{Cimrman, R.}, \bibinfo{author}{Henriksen, I.},
  \bibinfo{author}{Quintero, E.A.}, \bibinfo{author}{Harris, C.R.},
  \bibinfo{author}{Archibald, A.M.}, \bibinfo{author}{Ribeiro, A.H.},
  \bibinfo{author}{Pedregosa, F.}, \bibinfo{author}{{van Mulbregt}, P.}, and
  \bibinfo{author}{{SciPy 1.0 Contributors}} (\bibinfo{year}{2020}).
\newblock \bibinfo{title}{{SciPy 1.0: fundamental algorithms for scientific
  computing in Python}}.
\newblock \bibinfo{journal}{Nat. Methods} \textit{\bibinfo{volume}{17}},
  261--272.
\newblock \url{https://doi.org/10.1038/s41592-019-0686-2}.
\bibitem{sklearn2011}
\bibinfo{author}{Pedregosa, F.}, \bibinfo{author}{Varoquaux, G.},
  \bibinfo{author}{Gramfort, A.}, \bibinfo{author}{Michel, V.},
  \bibinfo{author}{Thirion, B.}, \bibinfo{author}{Grisel, O.},
  \bibinfo{author}{Blondel, M.}, \bibinfo{author}{Prettenhofer, P.},
  \bibinfo{author}{Weiss, R.}, \bibinfo{author}{Dubourg, V.},
  \bibinfo{author}{VanderPlas, J.}, \bibinfo{author}{Passos, A.},
  \bibinfo{author}{Cournapeau, D.}, \bibinfo{author}{Brucher, M.},
  \bibinfo{author}{Perrot, M.}, and \bibinfo{author}{Duchesnay, {\'E}.}
  (\bibinfo{year}{2011}).
\newblock \bibinfo{title}{{Scikit-learn: Machine Learning in Python}}.
\newblock \bibinfo{journal}{J. Mach. Learn. Res.}
  \textit{\bibinfo{volume}{12}}, 2825--2830.
\newblock \url{https://jmlr.org/papers/v12/pedregosa11a.html}.
\bibitem{statsmodels2010}
\bibinfo{author}{Seabold, S.}, and \bibinfo{author}{Perktold, J.}
  (\bibinfo{year}{2010}).
\newblock \bibinfo{title}{{Statsmodels: Econometric and Statistical Modeling
  with Python}}, in: \bibinfo{booktitle}{Proceedings of the 9th Python in
  Science Conference}, pp. \bibinfo{pages}{92--96}.
\newblock \url{https://doi.org/10.25080/Majora-92bf1922-011}.
\bibitem{silhouette}
\bibinfo{author}{Rousseeuw, P.J.} (\bibinfo{year}{1987}).
\newblock \bibinfo{title}{{Silhouettes: A graphical aid to the interpretation
  and validation of cluster analysis}}.
\newblock \bibinfo{journal}{J. Comput. Appl. Math.}
  \textit{\bibinfo{volume}{20}}, 53--65.
\newblock \url{https://doi.org/10.1016/0377-0427(87)90125-7}.
\bibitem{hog}
\bibinfo{author}{Dalal, N.}, and \bibinfo{author}{Triggs, B.}
  (\bibinfo{year}{2005}).
\newblock \bibinfo{title}{{Histograms of Oriented Gradients for Human
  Detection}}, in: \bibinfo{booktitle}{IEEE Computer Society Conference on
  Computer Vision and Pattern Recognition}, pp. \bibinfo{pages}{886--893}.
\newblock \url{https://doi.org/10.1109/cvpr.2005.177}.
\bibitem{hungarian}
\bibinfo{author}{Kuhn, H.W.} (\bibinfo{year}{1955}).
\newblock \bibinfo{title}{{The Hungarian method for the assignment problem}}.
\newblock \bibinfo{journal}{Nav. Res. Logist. Q.} \textit{\bibinfo{volume}{2}},
  83--97.
\newblock \url{https://doi.org/10.1002/nav.3800020109}.
\bibitem{otsu}
\bibinfo{author}{Otsu, N.} (\bibinfo{year}{1979}).
\newblock \bibinfo{title}{{A Threshold Selection Method from Gray-Level
  Histograms}}.
\newblock \bibinfo{journal}{IEEE Transactions on Systems, Man, and Cybernetics}
  \textit{\bibinfo{volume}{9}}, 62--66.
\newblock \url{https://doi.org/10.1109/TSMC.1979.4310076}.
\bibitem{mahalanobis1936}
\bibinfo{author}{Mahalanobis, P.C.} (\bibinfo{year}{1936}).
\newblock \bibinfo{title}{{On the generalised distance in statistics}}.
\newblock \bibinfo{journal}{Proceedings of the National Institute of Sciences
  of India} \textit{\bibinfo{volume}{2}}, 49--55.
\bibitem{ledoit_wolf}
\bibinfo{author}{Ledoit, O.}, and \bibinfo{author}{Wolf, M.}
  (\bibinfo{year}{2004}).
\newblock \bibinfo{title}{{A well-conditioned estimator for large-dimensional
  covariance matrices}}.
\newblock \bibinfo{journal}{Journal of Multivariate Analysis}
  \textit{\bibinfo{volume}{88}}, 365--411.
\newblock \url{https://doi.org/10.1016/S0047-259X(03)00096-4}.
\bibitem{icc}
\bibinfo{author}{Shrout, P.E.}, and \bibinfo{author}{Fleiss, J.L.}
  (\bibinfo{year}{1979}).
\newblock \bibinfo{title}{{Intraclass correlations: Uses in assessing rater
  reliability.}}
\newblock \bibinfo{journal}{Psychol. Bull.} \textit{\bibinfo{volume}{86}},
  420--428.
\newblock \url{https://doi.org/10.1037/0033-2909.86.2.420}.
\bibitem{ari}
\bibinfo{author}{Hubert, L.}, and \bibinfo{author}{Arabie, P.}
  (\bibinfo{year}{1985}).
\newblock \bibinfo{title}{{Comparing partitions}}.
\newblock \bibinfo{journal}{J. Classif.} \textit{\bibinfo{volume}{2}},
  193--218.
\newblock \url{https://doi.org/10.1007/bf01908075}.
\bibitem{nmi}
\bibinfo{author}{Vinh, N.X.}, \bibinfo{author}{Epps, J.}, and
  \bibinfo{author}{Bailey, J.} (\bibinfo{year}{2010}).
\newblock \bibinfo{title}{{Information Theoretic Measures for Clusterings
  Comparison: Variants, Properties, Normalization and Correction for Chance}}.
\newblock \bibinfo{journal}{J. Mach. Learn. Res.}
  \textit{\bibinfo{volume}{11}}, 2837--2854.
\newblock \url{https://jmlr.org/papers/v11/vinh10a.html}.
\bibitem{bcubed}
\bibinfo{author}{Amig{\'o}, E.}, \bibinfo{author}{Gonzalo, J.},
  \bibinfo{author}{Artiles, J.}, and \bibinfo{author}{Verdejo, F.}
  (\bibinfo{year}{2009}).
\newblock \bibinfo{title}{{A comparison of extrinsic clustering evaluation
  metrics based on formal constraints}}.
\newblock \bibinfo{journal}{Inf. Retr.} \textit{\bibinfo{volume}{12}},
  461--486.
\newblock \url{https://doi.org/10.1007/s10791-008-9066-8}.
\bibitem{bootstrap}
\bibinfo{author}{Efron, B.} (\bibinfo{year}{1979}).
\newblock \bibinfo{title}{{Bootstrap Methods: Another Look at the Jackknife}}.
\newblock \bibinfo{journal}{Ann. Stat.} \textit{\bibinfo{volume}{7}}, 1--26.
\newblock \url{https://doi.org/10.1214/aos/1176344552}.
\bibitem{mantel}
\bibinfo{author}{Mantel, N.} (\bibinfo{year}{1967}).
\newblock \bibinfo{title}{{The Detection of Disease Clustering and a
  Generalized Regression Approach}}.
\newblock \bibinfo{journal}{Cancer Res.} \textit{\bibinfo{volume}{27}},
  209--220.
\newblock \url{https://pubmed.ncbi.nlm.nih.gov/6018555/}.
\bibitem{quantile_regression}
\bibinfo{author}{Koenker, R.}, and \bibinfo{author}{Bassett, G.}
  (\bibinfo{year}{1978}).
\newblock \bibinfo{title}{{Regression Quantiles}}.
\newblock \bibinfo{journal}{Econometrica} \textit{\bibinfo{volume}{46}},
  33--50.
\newblock \url{https://doi.org/10.2307/1913643}.
\bibitem{holm}
\bibinfo{author}{Holm, S.} (\bibinfo{year}{1979}).
\newblock \bibinfo{title}{{A Simple Sequentially Rejective Multiple Test
  Procedure}}.
\newblock \bibinfo{journal}{Scand. J. Stat.} \textit{\bibinfo{volume}{6}},
  65--70.
\newblock \url{https://www.jstor.org/stable/4615733}.

\end{thebibliography}
\clearpage


\section*{Supplementary material}

Figure 1

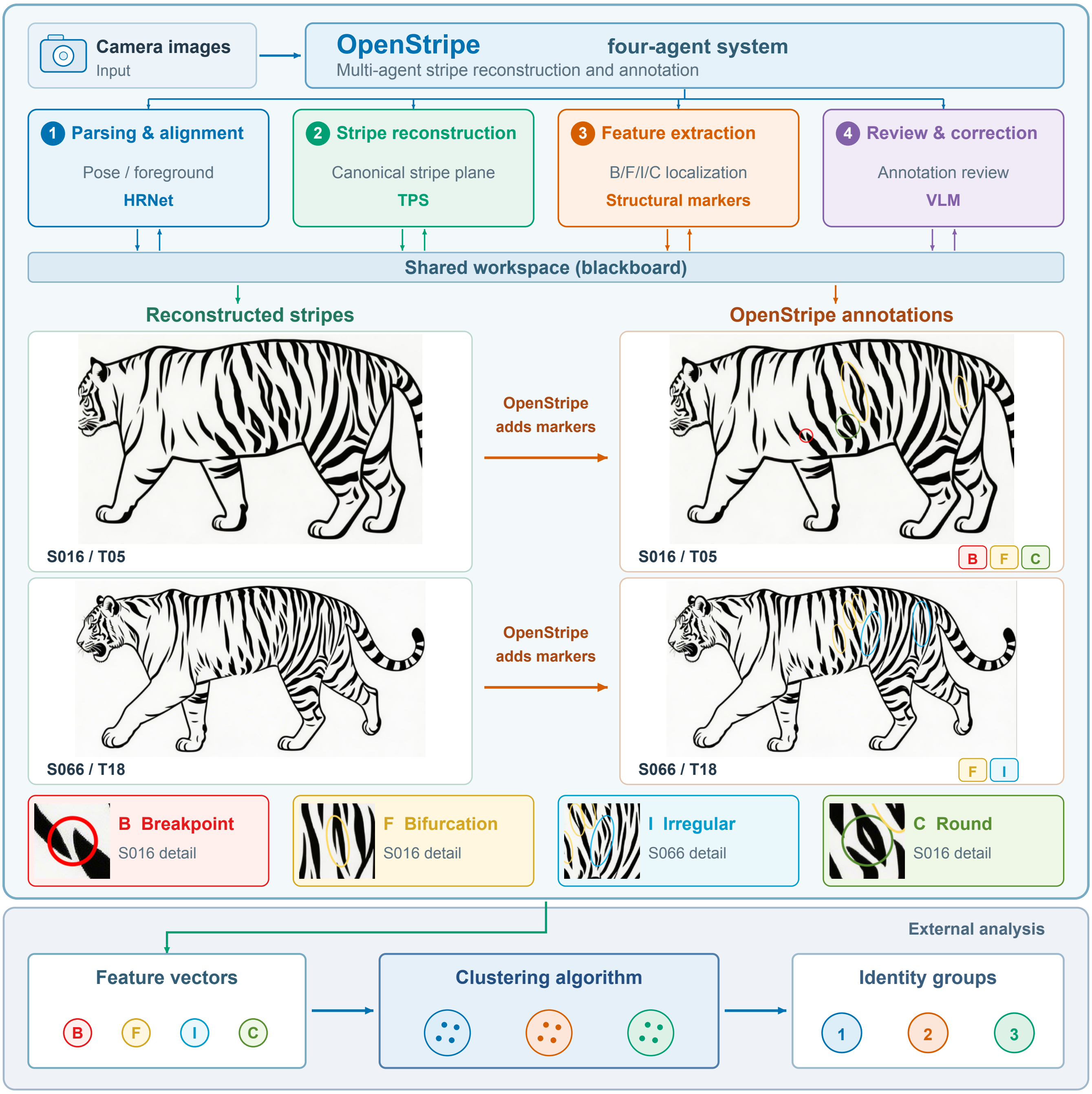

Camera images
Input
OpenStripe four-agent system
Multi-agent stripe reconstruction and annotation
1 Parsing & alignment
Pose / foreground
HRNet
2 Stripe reconstruction
Canonical stripe plane
TPS
3 Feature extraction
B/F/I/C localization
Structural markers
4 Review & correction
Annotation review
VLM
Shared workspace (blackboard)
Reconstructed stripes
OpenStripe annotations
S016 / T05
OpenStripe adds markers
S016 / T05
B F C
S066 / T18
OpenStripe adds markers
S066 / T18
F I
B Breakpoint
S016 detail
F Bifurcation
S016 detail
I Irregular
S066 detail
C Round
S016 detail
External analysis
Feature vectors
B F I C
Clustering algorithm
Identity groups
1 2 3

Figure 2

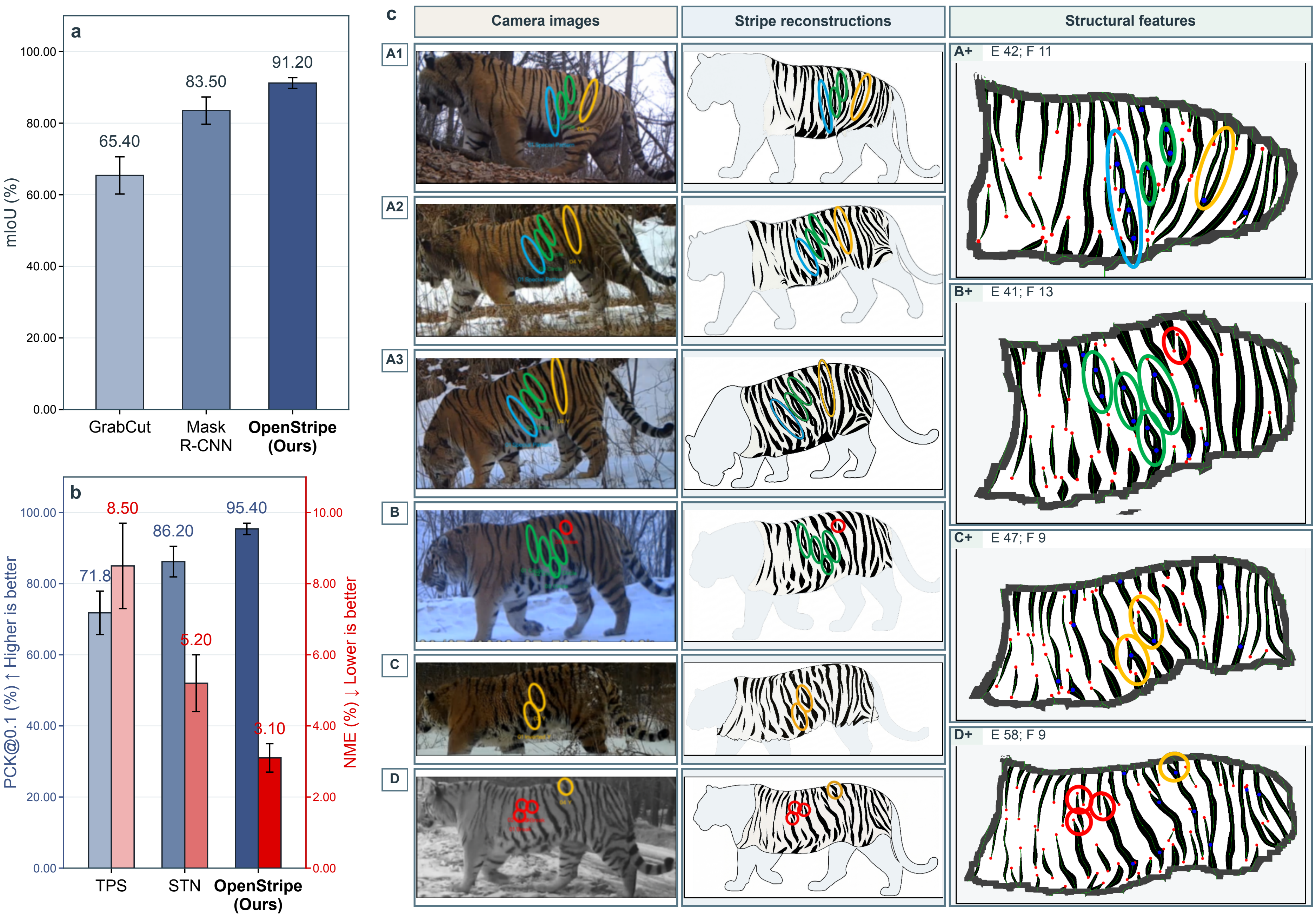

a
100.00
80.00
60.00
40.00
20.00
0.00
mIoU (%)
65.40
83.50
91.20
GrabCut
Mask R-CNN
OpenStripe (Ours)
b
PCK@0.1 (%) ↑ Higher is better
NME (%) ↓ Lower is better
10.00
8.00
6.00
4.00
2.00
0.00
71.8
8.50
86.20
5.20
95.40
3.10
TPS
STN
OpenStripe (Ours)
c
Camera images
Stripe reconstructions
Structural features
A1
A2
A3
B
C
D
A+
E 42; F 11
B+
E 41; F 13
C+
E 47; F 9
D+
E 58; F 9

## Figure 3

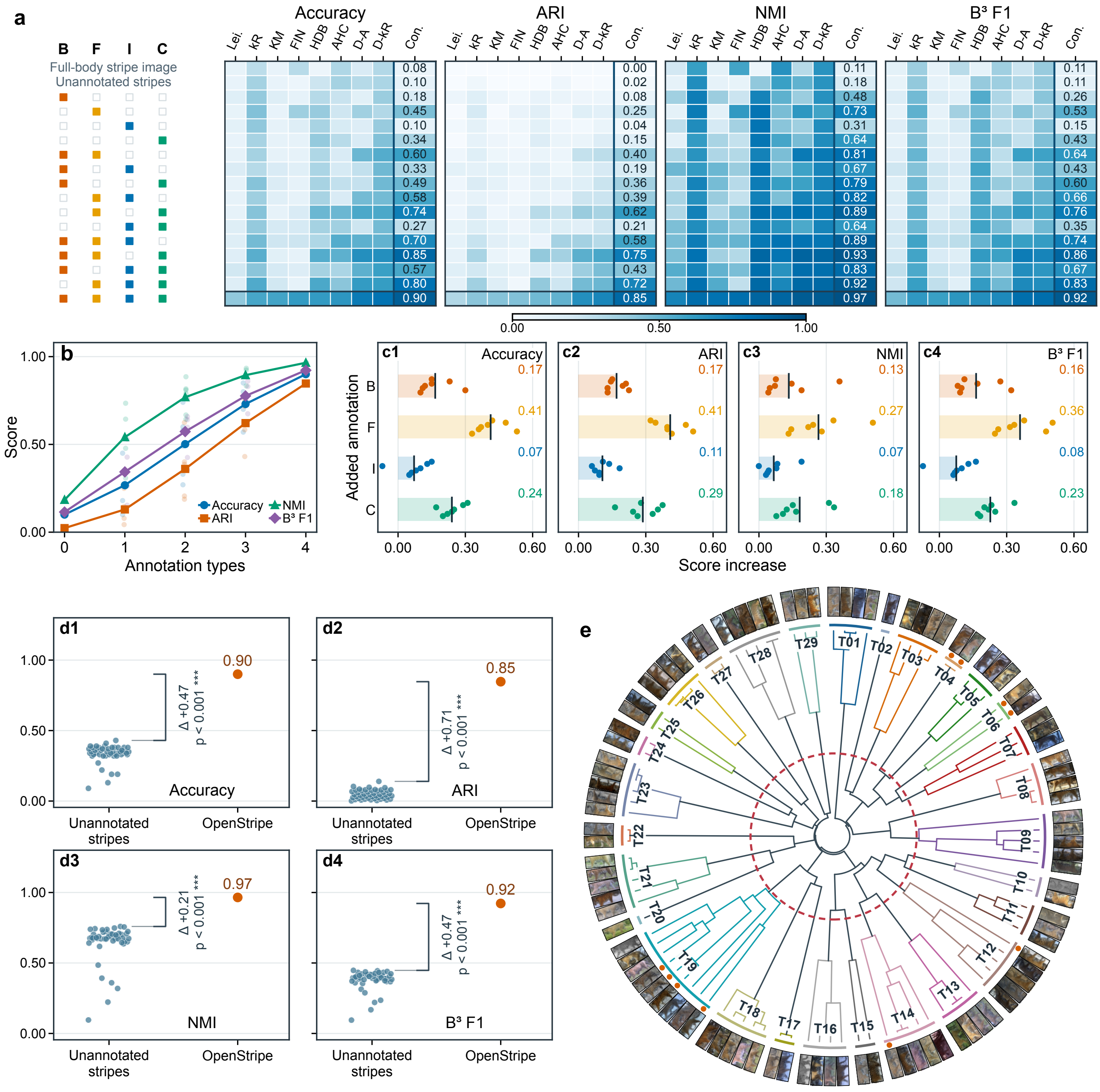

a
B F I C
Full-body stripe image
Unannotated stripes
Accuracy
ARI
NMI
B³ F1
Lei. kR KM FIN HDB AHC D-A D-kR Con.
b
Score
Annotation types
Accuracy
ARI
NMI
B³ F1
c1 Accuracy
c2 ARI
c3 NMI
c4 B³ F1
Added annotation
Score increase
d1 Accuracy
d2 ARI
d3 NMI
d4 B³ F1
Unannotated stripes
OpenStripe
e

**Figure 4**

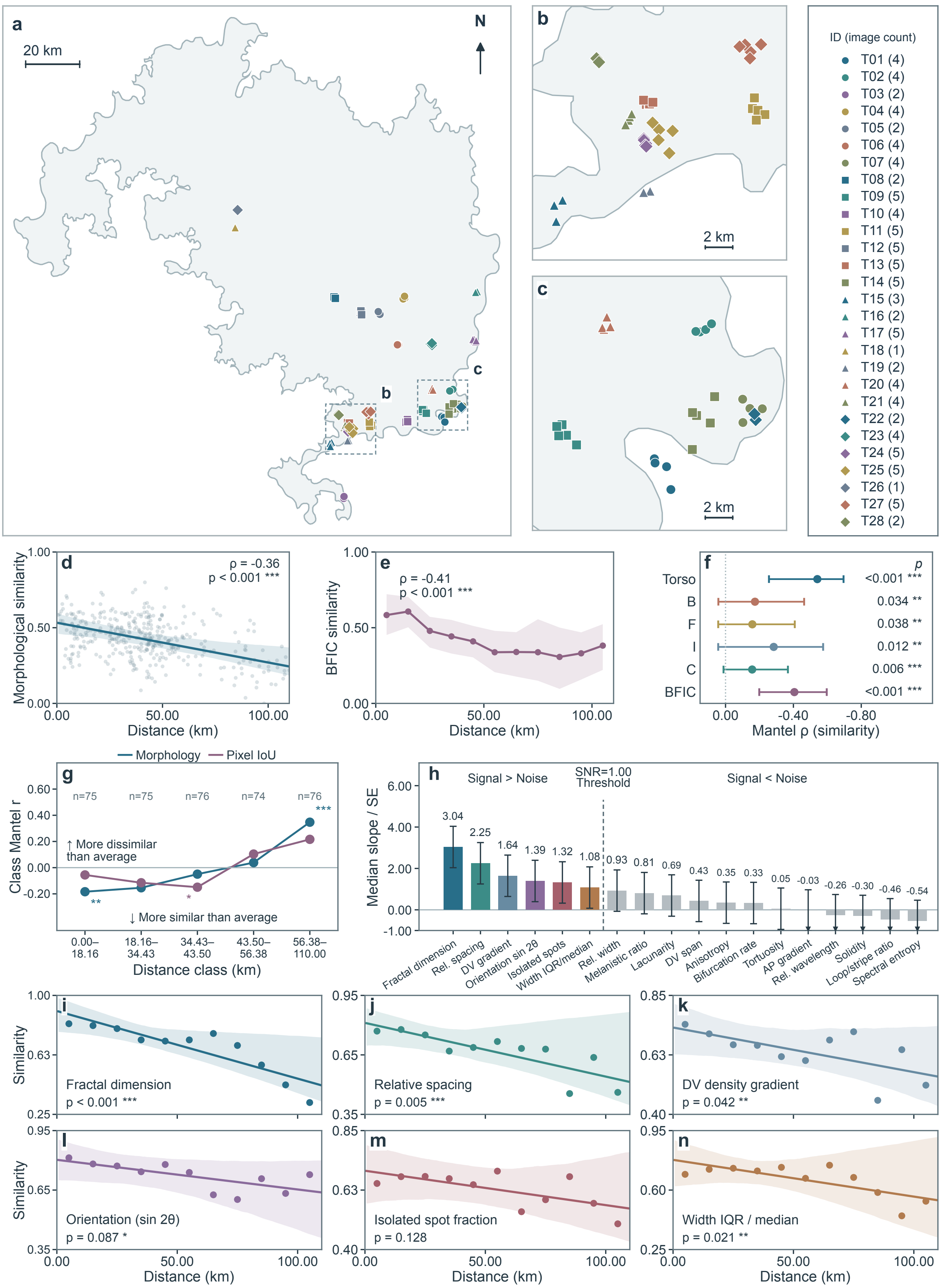
a
20 km
N
b
c
2 km
ID (image count)
T01 (4)
T02 (4)
T03 (2)
T04 (4)
T05 (2)
T06 (4)
T07 (4)
T08 (2)
T09 (5)
T10 (4)
T11 (5)
T12 (5)
T13 (5)
T14 (5)
T15 (3)
T16 (2)
T17 (5)
T18 (1)
T19 (2)
T20 (4)
T21 (4)
T22 (2)
T23 (4)
T24 (5)
T25 (5)
T26 (1)
T27 (5)
T28 (2)
d
ρ = -0.36
p < 0.001 ***
Morphological similarity
Distance (km)
e
ρ = -0.41
p < 0.001 ***
BFIC similarity
Distance (km)
f
Torso
B
F
I
C
BFIC
p
<0.001 ***
0.034 **
0.038 **
0.012 **
0.006 ***
<0.001 ***
Mantel ρ (similarity)
Morphology
Pixel IoU
g
n=75
n=75
n=76
n=74
n=76
↑ More dissimilar than average
↓ More similar than average
Class Mantel r
0.00–18.16
18.16–34.43
34.43–43.50
43.50–56.38
56.38–110.00
Distance class (km)
h
Signal > Noise
SNR=1.00 Threshold
Signal < Noise
Median slope / SE
3.04
2.25
1.64
1.39
1.32
1.08
0.93
0.81
0.69
0.43
0.35
0.33
0.05
-0.03
-0.26
-0.30
-0.46
-0.54
Fractal dimension
Rel. spacing
DV gradient
Orientation sin 2θ
Isolated spots
Width IQR/median
Rel. width
Melanistic ratio
Lacunarity
DV span
Anisotropy
Bifurcation rate
Tortuosity
AP gradient
Rel. wavelength
Solidity
Loop/stripe ratio
Spectral entropy
i
Fractal dimension
p < 0.001 ***
j
Relative spacing
p = 0.005 ***
k
DV density gradient
p = 0.042 **
l
Orientation (sin 2θ)
p = 0.087 *
m
Isolated spot fraction
p = 0.128
n
Width IQR / median
p = 0.021 **
Similarity
Distance (km)



\end{document}